\documentclass[11pt]{article}

\usepackage[preprint]{acl}
\usepackage{times}
\usepackage{latexsym}

\usepackage[T1]{fontenc}

\usepackage[utf8]{inputenc}

\usepackage{microtype}

\usepackage{inconsolata}

\usepackage{graphicx}

\usepackage{booktabs}
\usepackage{tabularx}
\usepackage{etoolbox}
\usepackage[table]{xcolor} 
\usepackage{nicematrix}
\usepackage{colortbl}
\usepackage{multirow}
\usepackage{amsmath}
\usepackage{amssymb}
\usepackage{amsfonts}

\title{SOAR: Source-Aware Reinforcement Learning with Precise Credit Assignment for Search Agents}

\author{
 \textbf{Shengze Huang\textsuperscript{1,2,*}},
 \textbf{Zihao Tang\textsuperscript{2,*}},
 \textbf{Jinluan Yang\textsuperscript{1,*}},
 \textbf{Xin Yu\textsuperscript{2,\textdaggerdbl}},
 \textbf{Zhenghao Lin\textsuperscript{2}},
\\
 \textbf{Xiao Liu\textsuperscript{2}},
 \textbf{Zengxuan Wen\textsuperscript{2}},
 \textbf{Yeyun Gong\textsuperscript{2}},
 \textbf{Haizhen Huang\textsuperscript{2}},
 \textbf{Weiwei Deng\textsuperscript{2}},
\\
 \textbf{Feng Sun\textsuperscript{2}},
 \textbf{Qi Zhang\textsuperscript{2}},
 \textbf{Fei Wu\textsuperscript{1}},
 \textbf{Kun Kuang\textsuperscript{1,\textdagger}}
\\
 \textsuperscript{1}Zhejiang University, \textsuperscript{2}Microsoft
\\
 \small{
   \textbf{Correspondence:} \href{mailto:kunkuang@cs.zju.edu.cn}{kunkuang@cs.zju.edu.cn}
 }
}
\begin{document}
\maketitle

\renewcommand{\thefootnote}{\fnsymbol{footnote}}
\footnotetext[1]{Equal contribution.}
\footnotetext[2]{Corresponding author.}
\footnotetext[3]{Project leader.}
\renewcommand{\thefootnote}{\arabic{footnote}}

\begin{abstract}
Reinforcement learning (RL) for search agents often suffers from evidence fabrication, a failure mode where agents bypass external information ($\mathcal{E}_{ext}$) and fabricates an incorrect answer. We argue that this fabrication arises from an over-reliance on unreliable internal knowledge ($\mathcal{E}_{int}$) as shortcuts due to the inherent opacity of reasoning trajectories. To address this, we propose \textbf{SOAR} (\textbf{SO}urce-\textbf{A}ware \textbf{R}einforcement learning), a paradigm that transforms monolithic reasoning into source-explicit trajectories. \textbf{SOAR} (i) disentangles $\mathcal{E}_{ext}$ and $\mathcal{E}_{int}$ into parallel, observable streams via a Source-Aware Workflow, and (ii) incorporates Source-Consistent Reward to guide the policy evolution. By making information-source attribution explicit, \textbf{SOAR} enables precise credit assignment between different knowledge streams. Evaluations show that \textbf{SOAR} outperforms baselines on 3B and 7B backbones by \textbf{7.3\%} and \textbf{5.4\%}, while reducing evidence fabrication by \textbf{74.8\%}. These confirm that SOAR significantly improves performance on knowledge-intensive benchmarks while effectively mitigating evidence fabrication, fostering more faithful and grounded reasoning with enhanced robustness.
\end{abstract}

\section{Introduction}

\begin{figure*}[t]
\centering

\includegraphics[width=\linewidth]{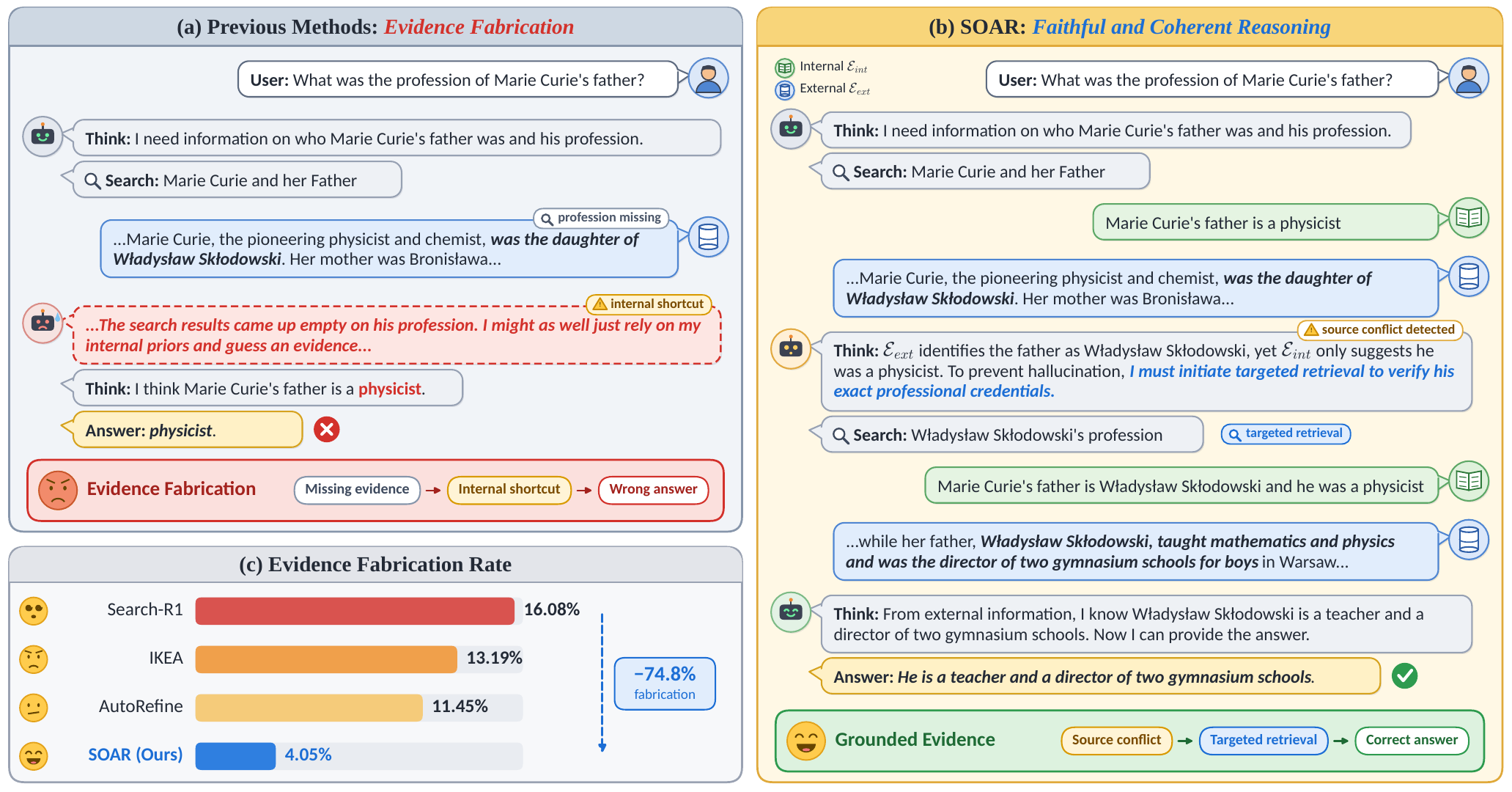}
\caption{Comparison of reasoning trajectories and evidence fabrication rates. (a) Previous methods often fall victim to evidence fabrication when initial retrieval results are incomplete, relying on hallucinated internal priors to fill gaps. (b) Our method (SOAR) implements a source-aware adjudication mechanism that monitors the alignment between internal knowledge ($\mathcal{E}_{int}$) and external evidence ($\mathcal{E}_{ext}$). By realizing the unreliability of internal information, the model initiates targeted retrieval to ensure a faithful reasoning trajectory. (c) Quantitative analysis shows SOAR reduces the evidence fabrication rate to 4.2\%, outperforming existing baselines.}
\label{fig:intro}
\end{figure*}

Large Language Models (LLMs) have demonstrated remarkable reasoning capabilities, yet they are inherently constrained by their static parametric knowledge~\cite{hendrycks2020measuring,wei2022chain,huang2023towards,schick2023toolformer, che2024hierarchical}. To mitigate hallucinations and handle knowledge-intensive tasks, recent works have evolved from static RAG to autonomous search agents~\cite{shi2025search, jin2025search, yao2022react, zheng2025deepresearcher}. By leveraging Reinforcement Learning (RL), these agents learn to interact with external search engines, iteratively retrieving evidence to support multi-step reasoning. Representative frameworks show that such "think-then-search" paradigms can significantly enhance performance in open-domain challenges.

Despite their promise, there exists a critical failure mode in search agent RL training: Evidence Fabrication, where the agent's reasoning evidence is not grounded in the retrieved information, as illustrated in Figure~\ref{fig:intro}(a). While previous studies~\cite{jia2025your,xu2025beyond,xinjie2025reagent} have also identified this phenomenon, they often attribute it to limited model capacity or data biases, overlooking that the architectural limitations of search agentic RL frameworks can themselves induce evidence fabrication. As quantified in Figure~\ref{fig:intro}(c), we measure the Evidence Fabrication Rate across prior approaches,with Search-R1~\cite{jin2025search} reaching 16.0\% , followed by IKEA~\cite{huang2025reinforced} at 13.4\% and AutoRefine~\cite{shi2025search} at 11.1\%, highlighting this open issue.

We argue that this phenomenon stems from an over-reliance on internal knowledge ($\mathcal{E}_{int}$) as shortcuts. This stems from the inherent opacity of monolithic reasoning trajectories developed in training, which fail to track whether the reasoning evidence comes from internal or external sources. This creates a credit assignment flaw: the reward signal cannot distinguish whether a successful outcome is a result of grounded synthesis of $\mathcal{E}_{ext}$ or an ungrounded heuristic from $\mathcal{E}_{int}$. Consequently, the RL process fails to provide precise signals needed to suppress internal shortcuts, ultimately reinforcing unfaithful reasoning behaviors instead of authentic evidence utilization.

To address this, we propose \textbf{SOAR} (\textbf{SO}urce-\textbf{A}ware \textbf{R}einforcement Learning). \textbf{SOAR} introduces two core innovations: (i) a \textbf{Source-Aware Workflow} that explicitly disentangles and organizes external information ($\mathcal{E}_{ext}$) and  internal knowledge ($\mathcal{E}_{int}$) into observable streams, and (ii) \textbf{Source-Consistent Reward } that enforces strict alignment between the model's self-declared evidence source and its reasoning process. \textbf{SOAR} makes the reasoning evidence fully traceable, and ensures precise credit assignment across trajectories with different evidence sources during training.

Our contributions are summarized as follows:
\begin{itemize}
    \setlength{\itemsep}{0pt}
    \setlength{\parskip}{0pt}
    \setlength{\parsep}{0pt}
    \item We propose \textbf{SOAR}, an RL framework with a \textbf{Source-Aware Workflow} that disentangles $\mathcal{E}_{ext}$ and $\mathcal{E}_{int}$ into observable streams to ensure precise credit assignment.
    \item We design a \textbf{Source-Consistent Reward} that enforces alignment between the model's self-declared sources and its reasoning process, preventing reliance on internal shortcuts.
    \item Evaluations across seven benchmarks show that SOAR consistently outperforms competitive RL baselines, achieving average relative improvements of \textbf{7.3\%} (3B) and \textbf{5.4\%} (7B). Notably, SOAR leads to a \textbf{74.81\%} relative reduction in evidence fabrication, demonstrating that our framework successfully guides the model toward faithful evidence-based reasoning instead of exploiting internal shortcuts.
\end{itemize}

\section{Preliminary}

\subsection{Search Agents}

Following prior work \cite{hu2018reinforcement, jin2025search, yao2022react, lewis2020retrieval}, we study search agents where an LLM policy $\pi_\theta$ interacts with external search tools to solve knowledge-intensive tasks. Given a query $x$, the agent produces a reasoning trajectory interleaved with search actions and retrieved observations, before generating the final answer $y$. Existing agentic RL approaches, typically based on GRPO-style optimization \cite{shao2024deepseekmathpushinglimitsmathematical}, optimize the policy using answer-level rewards:
\begin{equation}
\max_{\pi_\theta} \quad
\mathbb{E}_{x,y \sim \pi_\theta}
\left[r_\phi(x,y)\right]
-\beta D_{\mathrm{KL}}(\pi_\theta \,\|\, \pi_{\text{ref}}).
\end{equation}

However, such sparse supervision creates a credit assignment flaw: with the evidence sources hidden within the monolithic trajectories, the reward cannot distinguish whether a answer is derived from grounded external evidence or from a spurious reliance on internal knowledge.\textit{A comprehensive literature review on search agentic RL and faithful learning is detailed in Appendix~\ref{appendix:related_work}.}

\subsection{Internal vs. External Information}
We categorize information into two types: (i) \textbf{Internal Knowledge ($\mathcal{E}_{int}$)}, which comprises information implicitly stored within the model's parameters and accessible without tool interaction; and (ii) \textbf{External Information ($\mathcal{E}_{ext}$)}, representing task-specific information retrieved via search tools.

\subsection{Challenge}
\label{subsec:challenge}
We observe a systematic failure mode: \textbf{Evidence Fabrication}, as illustrated by the representative failure case in Figure~\ref{fig:intro}(a).

Evidence Fabrication refers to a pathological pattern where the agent's reasoning fundamentally departs from being grounded in the existing information context, particularly when the provided information lacks the necessary evidence to resolve the task. This tendency is systematically reflected in the elevated fabrication rates observed in conventional RL-based search agents (Figure~\ref{fig:intro}(c)).

This behavior poses a fundamental threat to the operational reliability of search agents, undermining both the utility of retrieved evidence and the agent's capacity for deep, persistent exploration when encountering complex tasks.

\begin{figure*}[h]
\centering
\includegraphics[width=\linewidth]{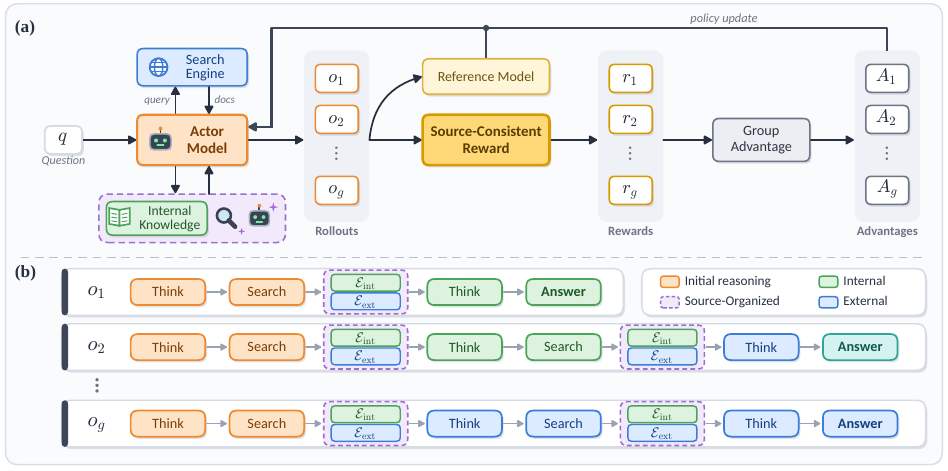}
\caption{\textbf{The training framework of SOAR.} (a) Overall workflow: for a given question $q$, the Actor Model interacts with a search engine and its internal knowledge base to generate a group of diverse reasoning trajectories $\{o_1, o_2, \dots, o_g\}$. These trajectories are evaluated by a reference model and rule-based rewards to compute the Group Advantage for reinforcement learning. (b) Fine-grained information selection: each trajectory $o_g$ represents a unique reasoning path derived from different selections and integration of $\mathcal{E}_{int}$ and $\mathcal{E}_{ext}$. By contrasting these trajectories, the Group Advantage facilitates precise credit assignment to specific information selection behaviors, guiding the agent to learn optimal strategies for grounding its final answer.}
\label{fig:training_scheme}
\end{figure*}

\section{Methodology}
\label{sec:method}

\subsection{Motivation}
While previous work might attribute evidence fabrication to limited model capacity, we argue that this pathological behavior is a consequence of over-reliance on internal knowledge ($\mathcal{E}_{int}$) as shortcuts. This stems from the inherent opacity of monolithic reasoning trajectories generated during training, which fail to track whether the reasoning evidence comes from internal or external sources. This leads to a \textbf{bidirectional credit assignment flaw} in search agentic RL: (i) Positive Conflation, where the same reward is directly assigned to both shortcut-driven and correct retrieval-augmented paths, which dilutes the gradient signal for proper grounding; and (ii) Negative Ambiguity, where the lack of detailed attribution prevents the model from receiving precise feedback, as the objective cannot distinguish between internal knowledge gaps and incorrect external evidence synthesis. Left unaddressed, this structural confusion encourages shortcut exploitation, causing the policy to fail to develop robust grounding mechanisms(see Appendix~\ref{sec:Case}).

To address this, we propose \textbf{SOAR} (\textbf{SO}urce-\textbf{A}ware \textbf{R}einforcement Learning), a paradigm that resolves the credit assignment flaw  by enforcing rigorous reasoning provenance tracing throughout the policy optimization process.

\subsection{Source-Aware Reinforcement Learning}

\subsubsection{Overview}
We formulate the learning process as a source-aware reinforcement learning problem (see Figure~\ref{fig:training_scheme}(a)). Building upon the GRPO framework, the policy $\pi_\theta$ is trained to maximize the expected reward within a structured observation space:

\begin{equation}
  \begin{split}
    \max_{\pi_\theta}\;
    \mathbb{E}_{q \sim \mathcal{D},\,
    y \sim \pi_\theta(\cdot \mid q,\mathcal{E}_{ext},\mathcal{E}_{int})}
    \!\left[ R_{sc}(q,y) \right]
    \\
    -\beta\, D_{\mathrm{KL}}(\pi_\theta \| \pi_{\mathrm{ref}})
  \end{split},
\end{equation}

where $\mathcal{E}_{ext}$ and $\mathcal{E}_{int}$ denote the external evidence and the elicited internal knowledge streams produced by the \textbf{source-aware workflow}. The \textbf{Source-Consistent Reward} $R_{sc}$ is designed to resolve the credit assignment ambiguity: it acts as a structural constraint that enforces strict alignment between each reasoning step and its self-declared evidence provenance. By conditioning the policy on these explicitly decoupled streams and optimizing $R_{sc}$, SOAR reconfigures the optimization landscape to incentivize a accurate and faithful policy.

\subsubsection{Source-Aware Workflow}
To resolve the lack of evidence source traceability, we reconstruct the reasoning process into \textbf{Source-Aware Workflow}. The core philosophy is to treat the model's internal parameters as a dynamic, query-contingent knowledge source that is accessed and recorded with the same structural rigor as an external corpus. This systematic organization makes the use of internal information explicit and aligns it with that of external information, effectively transforming the unconstrained utilization of both internal and external information into transparent, attributable agentic actions.

\paragraph{Context-Isolated Internal Elicitation} 
Within this workflow, for each atomic query $q_j$ generated, we initiate a dual-stream information gathering process. While a standard retriever $\mathcal{R}$ fetches external evidence $i_{ext}^{(j)}$ from a corpus $\mathcal{D}$, we simultaneously trigger a Pristine Self-Elicitation to manifest the model's internal state. 

To ensure the \textbf{functional independence} of internal recall and prevent it from prematurely contaminating the final answer, this elicitation is conducted within a \textit{Context-Isolated Environment}. We deliberately strip away the existing reasoning trajectory $\tau$, presenting only the specific query $q_j$ to the agent:
\begin{equation}
    i_{int}^{(j)} \sim P_{\theta}(\cdot \mid \text{Prompt}_{elic}, q_j),
\end{equation}
where $P_{\theta}$ denotes the probability distribution parameterized by the policy model, and $\text{Prompt}_{elic}$ is a fixed, neutral template. By restricting the conditioning set, we ensure that $i_{int}^{(j)}$ is a high-fidelity, explicit reflection of the model’s inherent parametric state, rather than a hallucination conditioned on the current reasoning context.

\paragraph{Source-Augmented Reasoning}
Once streams are elicited, the workflow integrates them into a unified yet structured source-aware observation, as showed in Figure~\ref{fig:training_scheme}(b). This symmetric presentation serves as a structural intervention that achieves two critical objectives for policy optimization:

\textbf{Elimination of Attribution Ambiguity:} By explicitly decoupling the conditioning space into $\mathcal{E}_{ext}$ and $\mathcal{E}_{int}$, we render the reasoning paths identifiable. This allows the reinforcement learning process to distinguish whether a successful trajectory is grounded in internal knowledge or external retrieved evidence, resolving the inherent untraceability of evidence sources in monolithic trajectories.

\textbf{Precise Credit Assignment via Source Selection:} Our framework facilitates an explicit adjudication between the elicited internal information and external evidence. During the GRPO process, the policy gradient acts as a selection mechanism that evaluates the \textbf{comparative reliability} of these sources. If outputs generated using a specific information stream consistently receive higher advantages $\hat{A}$ for a given query type, the policy learns to assign greater credit to that pathway. This enables the policy to dynamically navigate the trade-off between internal and external information, ultimately converging on a reasoning strategy that prioritizes the most robust source for each specific context.

Through this explicit decoupling and comparative credit assignment, \textbf{Source-Aware Workflow} ensures that learning signals are properly attributed to distinct pathways, enabling the principled optimization of information-acquisition strategies under reinforcement learning. The detailed prompt templates are provided in Appendix \ref{sec:appendix_prompts}.

\subsubsection{Source-Consistance Reward }
To effectively guide policy learning, we adopt a multi-dimensional reward function $R_{\text{total}} = R_{\text{ans}} \oplus R_{\text{fmt}} \oplus R_{\text{mat}}$. This formulation treats $R_{\text{fmt}}$ and $R_{\text{mat}}$ as \textbf{structural constraints} that shape the optimization landscape, ensuring that terminal accuracy ($R_{\text{ans}}$) is optimized within the boundaries of verifiable reasoning. Detailed reward definitions and coefficients are provided in Appendix~\ref{sec:reward_configuration}.

\paragraph{(i) Answer Correctness Reward ($R_{ans}$)}
Following \cite{jin2025empiricalstudyreinforcementlearning}, $R_{ans}$ provides the primary reinforcement signal for task completion. We assign a binary reward based on the exact match between the parsed answer and the ground truth $y_{gt}$:
\begin{equation}
R_{ans} = 
\begin{cases} 
1, & \text{if } \text{extract}(y) = y_{gt} \\
0, & \text{otherwise}
\end{cases}.
\end{equation}

\paragraph{(ii) Format Compliance Reward ($R_{fmt}$)}
To ensure the model adheres to the predefined agentic protocol (e.g., \texttt{<search>}, \texttt{<think>} tags), we implement a rule-based $R_{fmt}$. This serves as a formal constraint and a necessary condition for valid trajectories during RL training:
\begin{equation}
R_{fmt} = \alpha \cdot \mathbb{I}(y \text{ is structurally valid}).
\end{equation}

\paragraph{(iii) Source-Consistency Match Reward ($R_{mat}$)}
$R_{mat}$ enforces the faithfulness of the reasoning process by penalizing ambiguous outputs. It ensures a strict alignment between the agent's self-declared evidence source $C_y$ of its evidence stream $\mathcal{E}_{C_y}$:
\begin{equation}
R_{mat} = \gamma \cdot \mathbb{I}(\text{Verify}(y, \mathcal{E}_{C_y}) = 1),
\end{equation}
where $\text{Verify}(\cdot)$ evaluates whether the generated answer is consistent with the specified information stream. By default, We adopt an exact-match implementation for efficiency and compare it with an LLM-based verifier in Appendix~\ref{sec:verification_function}.

\begin{table*}[t]
\centering
\footnotesize
\setlength{\tabcolsep}{5pt}
\caption{Performance comparison of \textbf{SOAR} versus competitive baselines across diverse QA benchmarks. The results represent Exact Match (EM) accuracy on Qwen2.5-3B and 7B backbones. Bold denotes the best results. \textbf{Gray rows highlight our method and its relative improvement ($\Delta \%$) against the best-performing baseline.}}
\label{tab:main_results}
\begin{NiceTabular*}{\textwidth}{@{\extracolsep{\fill}}lcccccccc}[colortbl-like]
\toprule
& \multicolumn{3}{c}{\textbf{Single-Hop QA}} & \multicolumn{4}{c}{\textbf{Multi-Hop QA}} & \\
\cmidrule(lr){2-4} \cmidrule(lr){5-8}
\Block{1-1}{\textbf{Method}} & NQ & TriviaQA & PopQA & HotpotQA & 2Wiki & Musique & Bamboogle & \Block{1-1}{\textbf{Avg.}} \\
\midrule
\multicolumn{9}{c}{\textbf{Qwen2.5-3B-Base}} \\
\midrule
Direct Inference  & 0.106 & 0.288 & 0.108 & 0.149 & 0.244 & 0.020 & 0.024 & 0.134 \\
Naive RAG         & 0.348 & 0.544 & 0.387 & 0.255 & 0.226 & 0.047 & 0.080 & 0.270 \\
Search-R1-v0.3    & 0.393 & 0.593 & 0.445 & 0.359 & 0.369 & 0.160 & 0.289 & 0.372 \\
AutoRefine        & 0.453 & 0.619 & 0.447 & 0.341 & 0.366 & 0.135 & 0.341 & 0.386 \\
CoDA              & 0.403 & 0.578 & 0.460 & 0.364 & 0.273 & 0.133 & 0.344 & 0.365 \\
IKEA              & \textbf{0.464} & 0.625 & 0.469 & 0.316 & 0.342 & 0.131 & 0.325 & 0.382 \\
\rowcolor[gray]{0.9} \textbf{SOAR(Ours)} & 0.459 & \textbf{0.638} & \textbf{0.472} & \textbf{0.378} & \textbf{0.419} & \textbf{0.187} & \textbf{0.346} & \textbf{0.414} \\
\rowcolor[gray]{0.9} \quad $\Delta \%$ & $-1.0\%$ & \textbf{+2.1\%} & \textbf{+0.6\%} & \textbf{+3.8\%} & \textbf{+13.6\%} & \textbf{+16.9\%} & \textbf{+0.6\%} & \textbf{+7.3\%} \\
\midrule
\multicolumn{9}{c}{\textbf{Qwen2.5-7B-Base}} \\
\midrule
Direct Inference  & 0.134 & 0.408 & 0.140 & 0.183 & 0.250 & 0.031 & 0.120 & 0.181 \\
Naive RAG         & 0.349 & 0.585 & 0.392 & 0.299 & 0.235 & 0.058 & 0.208 & 0.304 \\
Search-R1-v0.3    & 0.461 & 0.636 & 0.447 & 0.412 & 0.363 & 0.180 & 0.411 & 0.415 \\
CoDA              & 0.422 & 0.661 & 0.456 & 0.416 & 0.359 & 0.158 & 0.320 & 0.399 \\
AutoRefine        & 0.472 & 0.660 & 0.435 & 0.428 & 0.378 & 0.157 & 0.446 & 0.425 \\
IKEA              & \textbf{0.482} & 0.659 & 0.478 & 0.368 & 0.363 & 0.146 & 0.319 & 0.402 \\
\rowcolor[gray]{0.9} \textbf{SOAR(Ours)} & 0.474 & \textbf{0.668} & \textbf{0.492} & \textbf{0.430} & \textbf{0.416} & \textbf{0.202} & \textbf{0.453} & \textbf{0.448} \\
\rowcolor[gray]{0.9} \quad $\Delta \%$ & $-1.7\%$ & \textbf{+1.1\%} & \textbf{+2.9\%} & \textbf{+0.5\%} & \textbf{+10.1\%} & \textbf{+12.2\%} & \textbf{+1.6\%} & \textbf{+5.4\%} \\
\midrule
\bottomrule
\end{NiceTabular*}
\end{table*}

\begin{table}[t] 
\centering
\caption{System prompt for Evidence Fabrication Judge.}
\label{tab:prompt_fab}
\footnotesize 
\begin{tabularx}{\linewidth}{X}
\toprule
\textbf{Task Instruction:} \\
Determine whether a model's reasoning is grounded in the provided information. Only judge whether the reasoning steps or the final answer actually depend on facts stated in the \texttt{<information>} block. Do NOT evaluate whether the final answer is correct.  \\
Output Format: \\
\texttt{\{} \\
\texttt{~~"reason": "...",} \\
\texttt{~~"grounded": true/false} \\
\texttt{\}} \\
\bottomrule
\end{tabularx}
\end{table}

\section{EXPERIMENTS}
\subsection{Experimental Setting}
\textbf{Datasets}
We evaluate our method on diverse single-hop and multi-hop QA benchmarks, including single-hop datasets: Natural Questions (NQ)~\cite{kwiatkowski2019natural}, TriviaQA~\cite{joshi2017triviaqa}, PopQA~\cite{mallen2023not}, and multi-hop datasets: HotpotQA~\cite{yang2018hotpotqa}, 2WikiMultihopQA (2Wiki)~\cite{ho2020constructing}, Musique~\cite{trivedi2022musique}, Bamboogle~\cite{press2023measuring}. Exact match (EM) accuracy is used as the primary evaluation metric across all datasets.We construct the training dataset by explicitly combining the training splits of NQ and HotpotQA., allowing our agent RL to learn both single-hop and multi-hop reasoning patterns.

\textbf{Baseline Methods} 
To assess the performance of our proposed SOAR framework, we select a comprehensive suite of baselines for comparative evaluation. Specifically, we include Direct Inference and Naive RAG, which are derived from Search-R1~\cite{jin2025search} (a work aligned with our experimental setup), alongside Search-R1~\cite{jin2025search, jin2025empiricalstudyreinforcementlearning},AutoRefine~\cite{shi2025search}, CoDA~\cite{liu2025coda}, IKEA~\cite{huang2025reinforced}, and AdaSearch~\cite{lin2025adasearch}. For fair comparison, we re-implement Search-R1, AutoRefine, CoDA, and IKEA via the publicly available code provided by their respective authors; the results of AdaSearch are adopted from its publication.

\textbf{Implementation Details} 
We use Qwen-2.5-3B/7B-Base~\cite{qwen2025qwen25technicalreport} as backbone agents, E5-base-v2~\cite{wang2022text} for retrieval, and the 2018 Wikipedia dump~\cite{karpukhin2020dense} as the corpus. The engine retrieves top-3 relevant documents per query. Models are trained via Group Relative Policy Optimization (GRPO).

\textit{Evaluation of Evidence Fabrication.} We employ GPT-5.6-Sol as the judge to assess evidence fabrication. We conduct the evaluation on the complete evaluation set and report results averaged over four independent runs. Table~\ref{tab:prompt_fab} presents the prompt template used for evidence fabrication evaluation.

\subsection{Main Results}
\label{sec:rq1}

We evaluate \textbf{SOAR} across diverse QA benchmarks, with additional Instruct-tuned results in Appendix~\ref{sec:Instruct-tuned Performance}. Table~\ref{tab:main_results} leads to the following findings:

\textbf{SOAR achieves state-of-the-art performance across benchmarks and scales.} 
As shown in Table~\ref{tab:main_results}, \textbf{SOAR} consistently outperforms competitive baselines on both 3B and 7B backbones. On the 3B scale, it reaches an average of 0.414, surpassing \textit{Search-R1} by +4.2\%. The gains are most pronounced in multi-hop reasoning: on Musique and 2Wiki, \textbf{SOAR} substantially outperforms monolithic-trajectory baselines, confirming that source transparency is critical for high-entropy, multi-step scenarios. Furthermore, the training dynamics for the 3B model, as illustrated in Figure~\ref{fig:combined_results}(a), show that the mean reward converges smoothly to a stable plateau.

\textbf{Consistent improvement with model scaling.} 
Performance scales reliably from 3B to 7B. On the 7B backbone, \textbf{SOAR} achieves an average of 0.448, outperforming \textit{AutoRefine} (0.425) and \textit{IKEA} (0.402). Notably, on Bamboogle it improves from 0.346 to 0.453, demonstrating that resolving the credit assignment flaw provides a strong foundation for scaling search agents.

\textbf{Faithfulness drives performance.} 
As shown in Figure~\ref{fig:intro}(c), \textbf{SOAR} reduces the Evidence Fabrication Rate to 4.05\% (a 74.81\% relative reduction) on the 3B backbone. In contrast to baselines that optimize for correctness without explicit grounding constraints, \textbf{SOAR} ensures that accuracy gains are driven by the effective selection and use of both internal and external knowledge, preventing the model from bypassing necessary retrieval with internal knowledge biases.

\textbf{Generalizability to harder QA.} 
\textbf{SOAR} also maintains clear advantages on the more challenging ASQA~\cite{stelmakh2022asqa} and WebWalkerQA~\cite{wu2025webwalker} datasets; results details are provided in Appendix~\ref{sec:appendix_hard_benchmarks}.

\subsection{Model Behaviors}
\label{sec:rq2}

\begin{figure*}[t] 
    \centering
    \includegraphics[width=\textwidth]{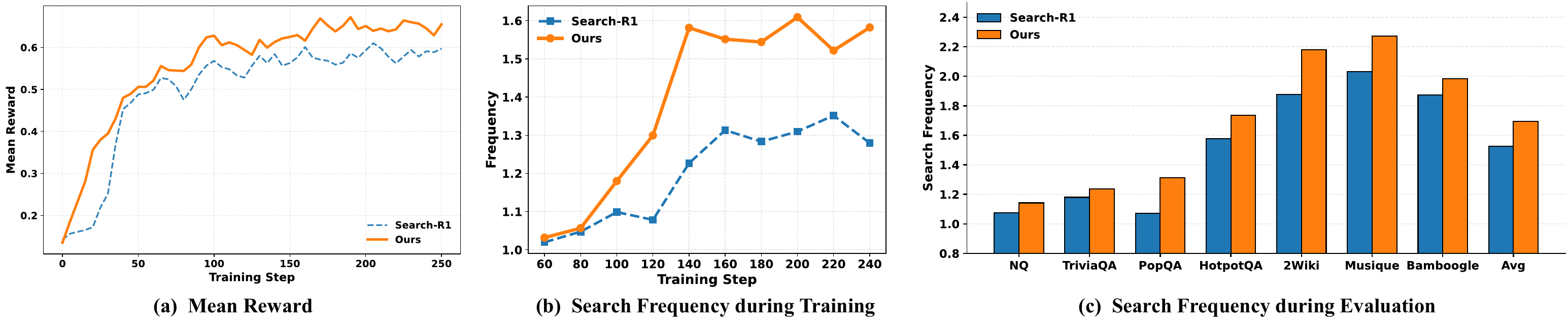} 
    \caption{Analysis of model behaviors during training and inference. (a) Mean reward convergence of SOAR and Search-R1 over training steps. (b) Search frequency dynamics during training: SOAR adaptively reduces evidence fabrication by shifting from internal shortcuts to proactive external search, as clearly reflected in the rising and stabilizing external retriever call rate. (c) Search frequency across benchmarks: the search-proactive policy learned during training generalizes consistently and robustly across diverse tasks.}
    \label{fig:combined_results}
\end{figure*}

\begin{table*}[h] 
\centering
\small
\setlength{\tabcolsep}{0pt} 
\caption{Ablation study on SOAR components. The “w/o Workflow \& Reward” row ablates both components. Removing the Source-Consistent Reward alone degrades performance; removing the Source-Aware Workflow additionally (i.e., removing both) leads to a further degradation. This confirms that both components are necessary.}
\label{tab:ablation_main}
\begin{tabularx}{\textwidth}{l @{\extracolsep{\fill}} ccccc} 
\toprule
& \multicolumn{4}{c}{\textbf{Multi-Hop QA}} & \\ 
\cmidrule(lr){2-5}
\multirow{-2}{*}{\textbf{Method}} & HotpotQA & 2Wiki & Musique & Bamboogle & \multirow{-2}{*}{\textbf{Avg}} \\
\midrule
\textbf{SOAR} & \textbf{0.378} & \textbf{0.419} & 0.187 & \textbf{0.346} & \textbf{0.333} \\
\hspace{1em} w/o Source-Consistent Reward & 0.364 & 0.388 & \textbf{0.192} & 0.333 & 0.319 \\
\hspace{1em} w/o Workflow \& Reward & 0.359 & 0.369 & 0.160 & 0.289 & 0.294 \\
\bottomrule
\end{tabularx}
\end{table*}

We investigate the efficacy of SOAR in optimizing the agent’s information-acquisition mechanism by resolving the \textbf{Credit Assignment Flaw}.

\textbf{Shift from Groundless Fabrication to Proactive Search.} 
As shown in Figure~\ref{fig:combined_results}(b), the search frequency consistently increases during the early stages of training and eventually stabilizes at a robust higher plateau (approximately 1.6 calls per query). By explicitly externalizing $\mathcal{E}_{int}$ through our \textbf{Source-Aware Workflow}, we expose the previously hidden shortcuts that lead to Evidence Fabrication. This structural transparency allows the \textbf{Source-Consistance Reward } to precisely penalize instances where the agent improperly relies on its internal knowledge at the expense of grounded reasoning. As the negative feedback loop suppresses these unreliable internal collapses, the agent is incentivized to seek more reliable evidence, resulting in a decisive strategic shift toward proactive search and a significantly more consistent external retrieval frequency as training progresses.

\textbf{Generalization of Search Behavior during Inference.} 
As illustrated in Figure~\ref{fig:combined_results}(c), this learned preference for external verification successfully generalizes across various evaluation benchmarks. Rather than being confined to the training distribution, \textbf{SOAR} consistently maintains an active search intensity across a wide spectrum of tasks, ranging from simple single-hop queries to complex multi-step reasoning problems. This cross-task consistency demonstrates that the search-proactive policy, which acquired by resolving the credit assignment flaw during training, effectively transfers to inference. This suggests that the model has internalized a source-aware decision mechanism that prioritizes evidentiary grounding over parametric heuristics when encountering unfamiliar knowledge boundaries. It ensures that the model’s reasoning remains anchored in verifiable evidence regardless of the specific evaluation context, confirming the generalizability of the information-acquisition behavior.

\begin{table*}[t]
\centering
\small
\setlength{\tabcolsep}{0pt} 
\caption{Ablation of reward learning on evidence fabrication rate across different answer groups.}
\vspace{-0.3\baselineskip}
\label{tab:ablation_fabrication}
\begin{tabularx}{\textwidth}{l @{\extracolsep{\fill}} cccccc} 
\toprule
& \multicolumn{2}{c}{\textbf{Answer-Correct}} & \multicolumn{2}{c}{\textbf{Answer-Wrong}} & \multicolumn{2}{c}{\textbf{Total}} \\
\cmidrule(lr){2-3} \cmidrule(lr){4-5} \cmidrule(lr){6-7}
\multirow{-2}{*}{\textbf{Method}} & Rate & Reduct. & Rate & Reduct. & Rate & Reduct. \\
\midrule
Search-R1           & 7.57\%  & — & 24.14\% & — & 16.08\% & — \\
\hspace{1em} w/o Source-Consistent Reward   & 2.75\%  & $\downarrow$63.67\% & 10.79\% & $\downarrow$55.30\% & 6.75\%  & $\downarrow$58.02\% \\
\textbf{SOAR}       & \textbf{1.34\%} & $\downarrow$82.30\% & \textbf{6.92\%} & $\downarrow$71.33\% & \textbf{4.05\%} & $\downarrow$74.81\% \\
\bottomrule
\end{tabularx}
\end{table*}

\subsection{Ablation Study}

We conduct comprehensive ablation experiments using the Qwen2.5-3B backbone to decompose the respective contributions of the \textit{Source-Aware Workflow} and \textit{Source-Consistent Reward}. The results are summarized in Table~\ref{tab:ablation_main} and Table~\ref{tab:ablation_fabrication}.

\textbf{Structural Transparency and Optimization Synergy.} 
As shown in Table~\ref{tab:ablation_main}, the full configuration of \textbf{SOAR} consistently achieves the highest performance across all multi-hop benchmarks. Removing both components (\textit{w/o reward \& workflow}) results in a significant performance decline from 0.333 to 0.294, confirming that structural transparency and precise optimization signals are both essential. Notably, while the workflow alone provides the structural transparency required to resolve source ambiguity, it is the integration of our source-consistent reward that pushes the model to its peak. This synergy suggests that the workflow establishes a navigable reasoning space, while the reward mechanism provides the necessary directional pressure to steer the policy toward evidence-grounded trajectories. This is particularly evident in challenging tasks, where the reward mechanism effectively reshapes the optimization landscape to exploit rigorous, evidence-based policy evolution.

\textbf{Reward as the Driver for Reasoning Faithfulness.} 
The impact of our reward shaping is even more pronounced in reducing evidence fabrication (Table~\ref{tab:ablation_fabrication}). While the workflow provides the observable streams necessary for source separation, the reward mechanism is the primary driver that compels the model to actually utilize them. By enforcing strict alignment between declarations and content, the reward mechanism further reduces the total fabrication rate from 6.75\% to 4.05\%. Crucially, even in \textit{Answer-Correct} instances, the reward nearly halves the fabrication rate from 2.75\% to 1.34\%. This indicates that the reward mechanism effectively penalizes spurious successes, where correct answers are bypassed as spurious hits if they rely on fabricated evidence. By suppressing these ungrounded reasoning paths, the reward shaping compels the agent to prioritize faithfulness over mere accuracy, ensuring that correct answers are derived from verifiable evidence rather than unreliable internal hallucinations.

\textbf{Token Efficiency.} 
We compare the per-turn generation cost between \textbf{SOAR} and the strongest baseline. \textbf{SOAR} maintains token efficiency comparable to baselines by parallelizing internal elicitation and external retrieval across GPUs. Detailed results and implementation analysis are in Appendix~\ref{sec:appendix_efficiency}.

\subsection{Systematic Robustness}
To evaluate the stability of policy learning, we conduct robustness experiments where the training environment is intentionally corrupted with noise.

\textbf{Experimental Setup.} We inject noise into 5\% of the training samples by targeting long-tail queries where the model's internal knowledge is sufficient. For these samples, we replace the correct retrieved evidence with misleading information to create a reasoning conflict. This setup evaluates whether the agent, when faced with localized contradictions representative of real-world retrieval noise, will mistakenly retreat into unreliable internal priors or maintain its learned search-proactive policy.

\textbf{Performance Resilience.} As shown in Table~\ref{tab:robustness_aligned}, \textbf{SOAR} exhibits superior robustness compared to baselines under adversarial pressure. While Search-R1 and IKEA suffer significant performance drops (averaging -0.038), \textbf{SOAR} maintains a stable performance with a decrease of only 0.019. This demonstrates that \textbf{SOAR} is less sensitive to noise, consistently identifying the optimal reasoning path even when specific training samples are corrupted.

\textbf{Faithfulness under Conflict.} Resilience of \textbf{SOAR} is further reflected in its reasoning integrity, as detailed in Table~\ref{tab:fabrication_robustness}. While noisy training drives baselines to distrust external sources, leading to a sharp increase in fabrication  rate (+2.15\% for \textit{Search-R1}), \textbf{SOAR}'s fabrication rate remains constant (+0.33\%). By resolving the credit assignment flaw, \textbf{SOAR} ensure the agent remains faithful to evidence despite environmental contradictions.

\textbf{Discussion.} These results confirm that \textbf{SOAR} effectively captures the dominant signal of high-quality retrieval. Specifically, in our adversarial setup, since the corrupted queries could be answered by internal knowledge, baselines are easily misled into over-relying on internal information. This pathological optimization leads to increased fabrication rates. Conversely, \textbf{SOAR}'s Source-Aware architecture ensures the agent maintains its search-proactive policy by filtering out noise and prioritizing the global grounding signal.

\begin{table}[t]
\centering
\small
\newcolumntype{Y}{>{\centering\arraybackslash}X}
\setlength{\tabcolsep}{3pt} 
\caption{QA Stability in Corrupted Training.}
\vspace{-0.3\baselineskip}
\label{tab:robustness_aligned}
\begin{tabularx}{\columnwidth}{l YYY}
\toprule
Method & Single-Hop & Multi-Hop & \textbf{Average} \\
\midrule
\textbf{SOAR (Ours)} & \textbf{0.545} & \textbf{0.375} & \textbf{0.448} \\
\hspace{1em} w/ disturb & \textbf{0.543} {\scriptsize($\downarrow$0.002)} & \textbf{0.345} {\scriptsize($\downarrow$0.030)} & \textbf{0.429} {\scriptsize($\downarrow$0.019)} \\
\midrule
Search-R1 & 0.515 & 0.342 & 0.415 \\
\hspace{1em} w/ disturb & 0.490 {\scriptsize($\downarrow$0.025)} & 0.290 {\scriptsize($\downarrow$0.052)} & 0.376 {\scriptsize($\downarrow$0.039)} \\
\midrule
AutoRefine & 0.522 & 0.352 & 0.425 \\
\hspace{1em} w/ disturb & 0.510 {\scriptsize($\downarrow$0.012)} & 0.305 {\scriptsize($\downarrow$0.047)} & 0.392 {\scriptsize($\downarrow$0.033)} \\
\midrule
IKEA & 0.540 & 0.299 & 0.402 \\
\hspace{1em} w/ disturb & 0.524 {\scriptsize($\downarrow$0.016)} & 0.246 {\scriptsize($\downarrow$0.053)} & 0.364 {\scriptsize($\downarrow$0.038)} \\
\bottomrule
\end{tabularx}

\end{table}
\begin{table}[t]
\centering
\small
\caption{Evidence Fabrication in Corrupted Training.}
\label{tab:fabrication_robustness}
\vspace{-0.3\baselineskip}
\newcolumntype{Y}{>{\centering\arraybackslash}X} 
\begin{tabularx}{\columnwidth}{l YYY} 
\toprule
Method & Default & w/ Disturb & \textbf{Increase} \\
\midrule
Search-R1 & 16.08\% & 18.23\% & +2.15\% \\
IKEA & 13.19\% & 14.98\% & +1.79\% \\
AutoRefine & 11.45\% & 12.96\% & +1.51\% \\
\midrule
\textbf{SOAR (Ours)} & \textbf{4.05\%} & \textbf{4.38\%} & \textbf{+0.33\%} \\
\bottomrule
\end{tabularx}
\end{table}
\section{Conclusion}

This work critically examines \textit{Evidence Fabrication} in search agents, identifying that \textbf{monolithic reasoning} and the resulting \textbf{credit assignment flaw} drive models to over-rely on internal shortcuts over external evidence. To address this, we propose \textbf{SOAR}, which replaces opaque reasoning with a \textbf{Source-Aware Workflow} to disentangle $\mathcal{E}_{int}$ and $\mathcal{E}_{ext}$ into observable streams. By leveraging this transparency, our \textbf{Source-Consistent Reward } resolves optimization ambiguity and enforces strict alignment between model claims and their underlying evidence. Extensive evaluations across seven benchmarks confirm that \textbf{SOAR} significantly improves performance while fostering a grounded information-acquisition policy, and does so at a controlled token cost relative to baselines. Ultimately, \textbf{SOAR} provides a scalable approach for anchoring agents in verifiable evidence from information stream, ensuring reasoning integrity during complex policy evolution.

\section{Limitations}

We discuss the current limitations of \textbf{SOAR} and outline directions for future work. 
Our primary training and evaluation rely on Wikipedia as a static, high-quality knowledge source, which enables clean diagnosis of the Evidence Fabrication issue but does not fully capture the extreme volatility of open-web environments. 
To address this, we have provided a preliminary validation on ASQA and WebWalkerQA in Appendix~\ref{sec:appendix_hard_benchmarks}, demonstrating that \textbf{SOAR} robustly maintains its advantage under online search settings. 
Nevertheless, testing the scaling limits of our source-aware mechanism under more severe retrieval constraints—such as handling actively malicious misinformation or highly unstructured, conflicting web sources—remains an important direction for future study.

\section{Ethics and Potential Risks}
The primary goal of this work is to enhance the truthfulness of search-based LLM agents. By modeling information sources, our approach reduces hallucinations and encourages evidence-grounded generation. All experiments utilize public academic benchmarks with no known personally identifiable information (PII). Despite these benefits, potential risks remain due to the reliance on external retrieval. Model outputs may inherit or amplify inaccuracies, biases, or incompleteness present in the underlying wiki corpus. Future work will explore more robust verification mechanisms to ensure the reliability of retrieved evidence.
\section{Acknowledgments}

\bibliography{ref}

\clearpage

\appendix
\section{Related Work}
\label{appendix:related_work}
\subsection{RL for Search Agents}
Reinforcement learning (RL) has been widely adopted to enhance the reasoning and decision-making capabilities of large language models (LLMs)~\cite{kaelbling1996reinforcement,jaech2024openai,guo2025deepseek, zheng2018drn}. By treating search calls as actions and optimizing outcome-based rewards, representative approaches such as Search-R1~\cite{jin2025search}, Search-o1~\cite{li2025search}, DeepResearcher~\cite{zheng2025deepresearcher}, Search-R1++~\cite{xu2026train}, Search-R2~\cite{he2026search} and IG-Search~\cite{liang2026ig} demonstrate that RL can effectively learn joint search-and-reasoning behaviors without explicit supervision over intermediate steps. Subsequent studies focus on improving reasoning ability through structured deployments or reward shaping that encourage the use of adaptive and economical tools~\cite{shi2025search,liu2025coda,liang2026othink}. Beyond general agents, several recent works like IKEA~\cite{huang2025reinforced}, AdaSearch~\cite{lin2025adasearch} and Dep-Search~\cite{liu2026dep}, explicitly study how to balance internal knowledge and external retrieval to improve efficiency by delineating knowledge boundaries. However, these methods primarily treat the use of internal versus external knowledge as a static decision or focus on search frequency. This leaves a blind spot where trajectories dominated by internal parametric knowledge may receive high rewards even when external evidence is ignored, encouraging shortcut behaviors and evidence fabrication.

In contrast, our work focuses on the dynamic selection and utilization of knowledge sources within the rollout process. By making knowledge-source attribution explicit, \textbf{SOAR} allows reinforcement learning to optimize over distinct knowledge-dependent decision paths rather than search efficiency alone, resolving the credit assignment flaw between internal and external sources.

\subsection{Faithful Learning in RL}
In LLMs RL-based post-training, shortcut behaviors are often manifested as factually inconsistent hallucinations or shallow generation patterns that correlate superficially with reward signals, rather than reflecting faithful reasoning~\cite{snyder2024early, tonmoy2024comprehensive,huang2025survey,lin2024flame,gui2026learning}. To alleviate these issues, prior work typically relies on output-level supervision and verification, such as preference modeling, reward verification, and self-consistency checks, which are usually implemented through external models or heuristic judges~\cite{ren2025knowrl,li2025hallucination,ding2025fapo,liu2025spark, xu2024face4rag,tamo2026evidencerl,zhang2026grad2reward}. In search agent RL scenarios, existing approaches to faithful learning in this context typically rely on external judges or verifier models to assess trajectory quality and evidence alignment~\cite{jia2025your,xu2025beyond,xinjie2025reagent,wen2026smartsearch}. While effective, such designs introduce additional training overhead and depend on the robustness of auxiliary models.

Our proposed \textbf{SOAR} is motivated by this critical and nuanced distinction. We address the issue by making knowledge-source attribution explicit within the rollout process, thereby transforming the monolithic reasoning process into a verifiable and structurally source-transparent trajectory without relying on external judges.

\begin{table*}[t] 
\centering 
\caption{Ablation of the Source-Consistency Match Reward: exact-match (EM) check vs.\ LLM judge (Claude Opus 4.8). Results are reported as mean $\pm$ std over 4 runs.} 
\label{tab:ablation-em-vs-judge} 
\resizebox{\textwidth}{!}{%
\begin{tabular}{lcccccccc} 
\toprule 
Method & NQ & TriviaQA & PopQA & HotpotQA & 2Wiki & Musique & Bamboogle & Avg \\ \midrule 
w/ EM check 
& $0.4642{\scriptstyle\pm0.0056}$ 
& $\mathbf{0.6370{\scriptstyle\pm0.0035}}$ 
& $0.4740{\scriptstyle\pm0.0043}$ 
& $\mathbf{0.3778{\scriptstyle\pm0.0039}}$ 
& $\mathbf{0.4168{\scriptstyle\pm0.0053}}$ 
& $0.1845{\scriptstyle\pm0.0048}$ 
& $0.3562{\scriptstyle\pm0.0073}$ 
& $0.4160{\scriptstyle\pm0.0016}$ \\ 
w/ Claude Opus 4.8 judge 
& $\mathbf{0.4698{\scriptstyle\pm0.0021}}$ 
& $0.6348{\scriptstyle\pm0.0046}$ 
& $\mathbf{0.4798{\scriptstyle\pm0.0046}}$ 
& $0.3692{\scriptstyle\pm0.0036}$ 
& $0.4142{\scriptstyle\pm0.0029}$ 
& $\mathbf{0.1917{\scriptstyle\pm0.0042}}$ 
& $\mathbf{0.3705{\scriptstyle\pm0.0091}}$ 
& $\mathbf{0.4187{\scriptstyle\pm0.0019}}$ \\ 
\bottomrule 
\end{tabular}%
} 
\end{table*}

\section{Implementation of Verification}
\label{sec:verification_function}

We implement $\text{Verify}(y, \mathcal{E}_{C_y})$ using a rule-based exact-match (EM) verifier for computational efficiency during RL training. To validate this design choice, we replace the EM verifier with an LLM-based verifier (Claude Opus 4.8) within the Source-Consistency Match Reward and repeat training under identical settings. 

As shown in Table~\ref{tab:ablation-em-vs-judge}, the LLM-based verifier yields only marginal improvements over EM verification (0.4187 vs.\ 0.4160 average performance), indicating that EM verification captures most source-consistency signals while avoiding the substantially higher token cost and inference latency of LLM-based verification. We therefore adopt EM verification throughout all experiments.

\section{Experimental Details}
\label{sec:appendix_experimental_details}
In this appendix, we provide additional implementations, dataset configurations, and environment setups to further ensure the reproducibility of our work. All revisions suggested through AI assistance were carefully reviewed and verified by the authors.

\subsection{Dataset Configurations and Statistics}
\label{sec:dataset_stats}
We evaluate our framework across multiple standard benchmarks. The detailed statistics of the datasets, including their specific splits and sample counts for both the training and evaluation phases, are comprehensively summarized in Table~\ref{tab:dataset_statistics}.

\begin{table}[ht]
\centering
\small
\renewcommand{\arraystretch}{1.2}
\caption{Detailed sample statistics and configurations of the datasets used for training and evaluation phases.}
\label{tab:dataset_statistics}

\begin{tabular}{@{}llc@{}}
\toprule
\textbf{Phase} & \textbf{Dataset} & \textbf{Number of Examples} \\
\midrule

\multirow{3}{*}{\textbf{Train}}
& HotpotQA & 90,447 \\
& Natural Questions (NQ) & 79,168 \\
& \textit{Total Train} & \textbf{169,615} \\

\midrule

\multirow{8}{*}{\textbf{Test}}
& PopQA & 14,267 \\
& 2WikiMultihopQA & 12,576 \\
& TriviaQA & 11,313 \\
& HotpotQA & 7,405 \\
& Natural Questions (NQ) & 3,610 \\
& MuSiQue & 2,417 \\
& Bamboogle & 125 \\
& \textit{Total Test} & \textbf{51,713} \\

\bottomrule
\end{tabular}

\end{table}
\subsection{Hyperparameters and Infrastructure}

Our training pipeline is implemented using the \texttt{veRL} framework. We employ the Group Relative Policy Optimization (GRPO) algorithm to fine-tune the Qwen2.5 model family (including Qwen2.5-3B, Qwen2.5-3B-Instruct, and Qwen2.5-7B) on a single node with 8$\times$ H200 NVIDIA GPUs. 

The training is performed for 3 epochs with a global batch size of 256 prompts per step. The actor network is optimized using a learning rate of $5 \times 10^{-7}$ with a linear warmup ratio of 0.285. For the GRPO policy rollout, we sample 5 rollout outputs per prompt at a temperature of 1.0 to compute group-relative rewards. The environment supports up to 5 search turns, utilizing an external retriever configured to fetch the top 3 relevant contexts.

\section{Reward Configuration}
\label{sec:reward_configuration}

Following Search-R1-v0.3, the final reward is computed as \begin{equation} R = R_{\text{ans}} + \alpha R_{\text{fmt}} + \gamma R_{\text{mat}}, \end{equation} where $R_{\text{ans}}$ denotes the answer correctness reward, $R_{\text{fmt}}$ denotes the format compliance reward, and $R_{\text{mat}}$ denotes the proposed Source-Consistency Match Reward. We set $\alpha=0.2$ and $\gamma=0.1$ in all experiments. The final reward is obtained by summing the three components and applying normalization.

\section{Performance on Instruct-tuned Backbone}
\label{sec:Instruct-tuned Performance}
To further verify the generalizability of \textbf{SOAR}, we evaluate its performance using the \textit{Qwen2.5-3B-Instruct} model as the backbone. As shown in Table~\ref{tab:instruct_results}, \textbf{SOAR} continues to demonstrate superior performance over competitive baselines. Notably, it achieves leading results in complex multi-hop tasks like \textit{2Wiki} (0.411) and \textit{Musique} (0.144), resulting in a leading average score of \textbf{0.397}.

\begin{table*}[t]
\centering
\small
\setlength{\tabcolsep}{10pt}
\caption{Performance comparison on the \textbf{Qwen2.5-3B-Instruct} backbone. Bold denotes the best results.}
\label{tab:instruct_results}
\begin{tabular*}{\textwidth}{@{\extracolsep{\fill}}lcccccccc}
\toprule
& \multicolumn{3}{c}{\textbf{Single-Hop QA}} & \multicolumn{4}{c}{\textbf{Multi-Hop QA}} & \\
\cmidrule(lr){2-4} \cmidrule(lr){5-8}
\multirow{-2}{*}{Method} & NQ & TriviaQA & PopQA & HotpotQA & 2Wiki & Musique & Bamboogle & \multirow{-2}{*}{Avg} \\
\midrule
\multicolumn{9}{c}{\textbf{Qwen2.5-3B-Instruct}} \\
\midrule
Search-R1-v0.3   & 0.350 & 0.572 & 0.395 & 0.301 & 0.318 & 0.112 & 0.228 & 0.325 \\
AutoRefine       & 0.424 & 0.587 & 0.400 & 0.348 & 0.364 & 0.144 & 0.320 & 0.369 \\
AdaSearch        & 0.379 & 0.568 & 0.428 & 0.334 & 0.385 & 0.144 & 0.280 & 0.360 \\
IKEA             & 0.444 & 0.632 & \textbf{0.465} & 0.329 & 0.347 & 0.125 & 0.276 & 0.375 \\
\textbf{Ours}    & \textbf{0.451} & \textbf{0.636} & 0.452 & \textbf{0.368} & \textbf{0.411} & \textbf{0.144} & \textbf{0.325} & \textbf{0.397} \\
\bottomrule
\end{tabular*}
\end{table*}

\section{Performance on Challenge benchmark}
\label{sec:appendix_hard_benchmarks}

To further evaluate \textbf{SOAR} under more complex conditions, we conduct experiments on ASQA and WebWalkerQA, two benchmarks that require synthesizing information from multiple documents and involve longer reasoning chains.

\textbf{Experimental Setup.}
For ASQA, we use E5-base-v2 as the retriever, with the 2018 Wikipedia dump as the corpus, retrieving the top-3 relevant documents per query. For WebWalkerQA, we migrate \textbf{SOAR} to an online search setting, where the agent queries a live search API and receives the top-3 returned results per query, mirroring the retrieval budget of the offline setting.

\textbf{Results.}
As shown in Table~\ref{tab:hard_benchmarks}, \textbf{SOAR} consistently outperforms competitive baseline on both benchmarks. The performance advantage observed in the main experiments transfers robustly to these harder, long-form QA settings, confirming that the workflow generalizes effectively to real-world online search and complex reasoning scenarios.

\begin{table}[ht]
\centering
\small
\setlength{\tabcolsep}{5pt}
\caption{Zero-shot performance comparison on challenging long-form QA benchmarks using the Qwen2.5-7B backbone. Evaluation metrics include F1 score for ASQA and Accuracy judged by GPT-4o-mini for WebWalkerQA. Bold text indicates the best performance.}
\label{tab:hard_benchmarks}
\begin{tabular}{@{}lccccc@{}}
\toprule
& \textbf{ASQA} & \multicolumn{4}{c}{\textbf{WebWalkerQA}} \\
\cmidrule(lr){2-2} \cmidrule(lr){3-6}
\textbf{Method} & F1 & Easy & Mid & Hard & Avg \\
\midrule
Search-R1 & 66.5 & 27.5 & 20.0 & 17.9 & 21.1 \\
\textbf{SOAR} & \textbf{68.5} & \textbf{29.4} & \textbf{26.1} & \textbf{21.3} & \textbf{25.2} \\
\bottomrule
\end{tabular}
\end{table}

\section{Additional Analysis}
\subsection{Token Cost Analysis}
\label{sec:appendix_efficiency}

We compare per-turn generation efficiency between \textbf{SOAR} and the strongest baseline. As shown in Table~\ref{tab:token_cost}, Our method produces an average of 431 output tokens per turn, compared to 408 tokens for the baseline Search-R1. The slight increase stems from the explicit internal knowledge elicitation step, and we mitigate this overhead through two design choices. First, we strictly control the length and quantity of internal knowledge: all noisy context is removed and only the essential entities are retained, substantially reducing the tokens cost in this step. Second, internal elicitation and external retrieval are executed in parallel; moreover, we leverage the veRL framework to generate internal knowledge concurrently across GPUs. These optimizations ensure that \textbf{SOAR} achieves its performance gains with only a marginal increase in token cost and no additional peak VRAM usage relative to the baseline.

\begin{table}[ht]
\centering
\caption{Performance and efficiency trade-off on 3B backbone. \textit{Acc} denotes the average score across the 7 main benchmarks. \textit{Token Cost} refers to the average output tokens per reasoning turn, encompassing both information and thinking content.}
\label{tab:token_cost}
\begin{tabular}{@{}lcc@{}}
\toprule
\textbf{Method} & \textbf{Acc} & \textbf{Token Cost} \\
\midrule
Search-R1 & 0.372 & 408.4 \\
AutoRefine & 0.386 & 415.9 \\
CoDA & 0.365 & 745.2 \\
IKEA & 0.382 & \textbf{396.7} \\
\textbf{SOAR} & \textbf{0.414} & 431.0 \\
\bottomrule
\end{tabular}
\end{table}

\subsection{Statistical Significance Analysis}
\label{app:significance}

To verify that the improvements of SOAR are statistically reliable rather than artifacts of sampling noise, we run inference for both Search-R1 and SOAR under four random seeds with a sampling temperature of $1.0$, and report the mean and standard deviation in Table~\ref{tab:seed-variance}. The run-to-run variance is consistently small for both methods (average standard deviation $\leq 0.0021$). In contrast, SOAR improves over Search-R1 by $+4.74$ points on the average score ($0.4187$ vs.\ $0.3713$), which is more than an order of magnitude larger than the observed variance. Moreover, SOAR outperforms Search-R1 on every benchmark across all seeds. These results confirm that the gains of SOAR are stable and consistent.

\begin{table*}[t]
\centering
\caption{Performance of Search-R1 and SOAR under four random seeds (temperature $=1.0$) on the Qwen2.5-3B backbone. Results are reported as mean~$\pm$~std over 4 runs. Bold denotes the best results.}
\label{tab:seed-variance}
\resizebox{\textwidth}{!}{%
\begin{tabular}{lcccccccc}
\toprule
Method & NQ & TriviaQA & PopQA & HotpotQA & 2Wiki & Musique & Bamboogle & Avg \\
\midrule
Search-R1
& $0.3902{\scriptstyle\pm0.0031}$
& $0.5940{\scriptstyle\pm0.0029}$
& $0.4412{\scriptstyle\pm0.0041}$
& $0.3582{\scriptstyle\pm0.0017}$
& $0.3715{\scriptstyle\pm0.0031}$
& $0.1590{\scriptstyle\pm0.0014}$
& $0.2850{\scriptstyle\pm0.0086}$
& $0.3713{\scriptstyle\pm0.0021}$ \\
SOAR
& $\mathbf{0.4698{\scriptstyle\pm0.0021}}$
& $\mathbf{0.6348{\scriptstyle\pm0.0046}}$
& $\mathbf{0.4798{\scriptstyle\pm0.0046}}$
& $\mathbf{0.3692{\scriptstyle\pm0.0036}}$
& $\mathbf{0.4142{\scriptstyle\pm0.0029}}$
& $\mathbf{0.1917{\scriptstyle\pm0.0042}}$
& $\mathbf{0.3705{\scriptstyle\pm0.0091}}$
& $\mathbf{0.4187{\scriptstyle\pm0.0019}}$ \\
\bottomrule
\end{tabular}%
}
\end{table*}

\subsection{Human Verification of Fabrication Evaluation}
\label{app:human-verification}

We conduct a human verification to examine the reliability of the automatic evaluation. Specifically, we randomly sample 200 judge decisions from the evaluation outputs and manually verify whether each decision correctly identifies the presence or absence of evidence fabrication according to the same criteria as the judge prompt. The judge agrees with human annotation in 98.5\% of the cases (197 out of 200), indicating that the automatic fabrication evaluation is highly consistent with human judgment and that the reported fabrication rates are reliable.

\section{Agent Prompt Templates}
\label{sec:appendix_prompts}

\subsection{Agent RL Training Prompt}

Table~\ref{tab:prompt_rl} details the system prompt used for Agent RL training, which guides the agent's interaction with the external environment during the reinforcement learning process.

\begin{table*}[t]
\centering
\caption{System prompt used for Agent RL training.}
\label{tab:prompt_rl}
\small 
\begin{tabularx}{0.95\linewidth}{X}
\toprule
\midrule
You are an expert assistant capable of solving knowledge-intensive tasks efficiently. You will be given a question to answer as accurately as possible by carefully balancing \textbf{INTERNAL} information and \textbf{EXTERNAL} information. \\
\addlinespace[0.3em]
If you find you lack some knowledge, you can call an information engine by \texttt{\textless search\textgreater\ query \textless/search\textgreater}. It will return the top internal and external information. \\
\addlinespace[0.3em]
To arrive at the answer, you will proceed step-by-step in a structured cycle of \texttt{\textless think\textgreater}, \texttt{\textless search\textgreater}, \texttt{\textless information\textgreater}, and \texttt{\textless information\_source\textgreater} sequences. At end, you should use \texttt{\textless answer\textgreater} to provide answer. You can only generate content within these special tags. \\
\addlinespace[0.3em]
\textbf{Thinking Phase (\texttt{\textless think\textgreater}):} For question, it may be decomposed into sub-questions for you to think about. Some sub-questions may be answered by searching, while others may not. You can also use the \texttt{\textless think\textgreater} tag to express your uncertainty about the sub-question, weigh the accuracy and usefulness of both types of information, and decide whether further search is needed. \\
\addlinespace[0.3em]
\textbf{Searching Phase (\texttt{\textless search\textgreater}):} Formulate a search query only if required to fill a knowledge gap or verify uncertainty. \\
\addlinespace[0.3em]
\textbf{Information Phase (\texttt{\textless information\textgreater}):} Use internal and external results as context for further steps. \\
\addlinespace[0.3em]
\textbf{Information Source (\texttt{\textless information\_source\textgreater}):} Before providing your final answer, indicate clearly whether you are using your own internal knowledge or external information. Only output \textbf{either ``internal'' or ``external''} inside this tag, nothing else. \\
\addlinespace[0.3em]
\textbf{Answering Phase (\texttt{\textless answer\textgreater}):} Provide a concise and accurate answer within tags once you have enough knowledge. The answer should be short and precise, such as \texttt{\textless answer\textgreater\ Beijing \textless/answer\textgreater}. \\
\addlinespace[0.3em]
\textbf{Here are some examples:} \\
\addlinespace[0.3em]
Example 1: search is needed, search more than once. \\
Question: xxx \\
\textless think\textgreater\ xxx \textless/think\textgreater\ \textless search\textgreater\ xxx \textless/search\textgreater\ \textless information\textgreater\ internal: xxx external: xxx \textless/information\textgreater\ ... (search more than once) \textless information\_source\textgreater\ internal/external \textless/information\_source\textgreater\ \textless answer\textgreater\ xx \textless/answer\textgreater \\
\addlinespace[0.3em]
Example 2: search is needed, only search once. \\
Question: xxx? \\
\textless think\textgreater\ xxx \textless/think\textgreater\ \textless search\textgreater\ xxx \textless/search\textgreater\ \textless information\textgreater\ internal: xxx external: xxx \textless/information\textgreater\ \textless information\_source\textgreater\ internal/external \textless/information\_source\textgreater\ \textless answer\textgreater\ xx \textless/answer\textgreater \\
\addlinespace[0.3em]
Each search should be focused on one sub-question. The answer within \texttt{\textless answer\textgreater} tags should be short and precise, such as \texttt{\textless answer\textgreater\ yes \textless/answer\textgreater}. \\
\addlinespace[0.3em]
Now it is your turn to answer the question. \\
\textbf{Question:} \{question\} \\
\bottomrule
\end{tabularx}
\end{table*}

\subsection{Internal Knowledge Extraction Prompt}
Table~\ref{tab:prompt_internal} presents the prompt template used to extract $\mathcal{E}_{int}$. This prompt enforces a strict format to ensure that the internal knowledge is captured as a set of $k$ independent, declarative facts, providing a structured basis for the subsequent source-consistency evaluation.

\begin{table*}[t]
\centering
\caption{System prompt used for Internal Knowledge Extraction.}
\label{tab:prompt_internal}
\small 
\begin{tabularx}{0.95\linewidth}{X}
\toprule
\midrule
You are a helpful assistant. \\
\addlinespace[0.3em]
Your task is to list knowledge relevant to the question using \textbf{ONLY} your internal knowledge. Do \textbf{NOT} invent, guess, speculate, or hallucinate any facts. If your internal knowledge is limited, state only what is confidently known. \\
\addlinespace[0.3em]

\textbf{You must STRICTLY follow all rules below:} \\
\addlinespace[0.3em]
1) Your output \textbf{MUST} contain exactly \textbf{\texttt{\{top-$k$\}}} items. \\
2) Each item \textbf{MUST} be a single declarative sentence. \\
3) Each item \textbf{MUST} be formatted as a numbered bullet. \\
4) Output \textbf{ONLY} the \texttt{\{top-$k$\}} numbered bullets and nothing else. \\
5) Do \textbf{NOT} include explanations, reasoning, summaries, or meta commentary. \\
\bottomrule
\end{tabularx}
\end{table*}

\section{Case study}
\label{sec:Case}
\subsection{Spurious Correctness from evidence fabrication}
To further analyze model behaviors beyond final-answer accuracy, we examine cases where the baseline(Search-R1) produces the correct answer despite failing to retrieve sufficient supporting evidence. Table~\ref{tab:case_study_1} presents a representative example.

We observe that \textbf{Search-r1} occasionally reaches the correct prediction without retrieving key evidence required for solving the question. Instead, the model fabricates evidence to fill missing information. Although the final answer happens to be correct, the reasoning process is not fully grounded in retrieved evidence, revealing a hidden retrieval failure. Such behaviors may also undermine reinforcement learning optimization, as correct predictions supported by unreliable reasoning can receive positive rewards, potentially reinforcing suboptimal retrieval patterns.

In contrast, \textbf{SOAR} continues retrieval when evidence remains incomplete, encouraging evidence-grounded reasoning instead of correctness by chance. This behavior improves the reliability of knowledge-intensive reasoning beyond final-answer accuracy.

\subsection{Source-Aware Reasoning Improves Deep Evidence Retrieval}
To further investigate the effectiveness of SOAR in complex scenarios, we conduct a case study on a multi-hop factual question: \textit{"Which film was released earlier, Ek Ladka Ek Ladki or The Original Latin Kings Of Comedy?"} This task is challenging as it requires the model to aggregate and compare temporal data from disparate sources.

Table \ref{tab:case_study_2} highlights a typical failure mode of \textbf{Search-r1}, where the model resorts to unreliable hallucinations when faced with insufficient external evidence. The initial search only retrieves information for one film. Instead of seeking the missing data, the model relies on its internal information, leading to a hallucinated release date (1978) and an incorrect final answer.

In contrast, \textbf{SOAR} demonstrates superior \textbf{self-verification capabilities}. Upon identifying that the retrieved information is incomplete and that its internal knowledge might be unreliable, SOAR proactively initiates a second search targeted at the missing entity. By synthesizing the confirmed dates from multiple external sources (1992 vs. 2002), it successfully resolves the conflict and derives the correct conclusion. This case highlights how SOAR effectively mitigates hallucinations in knowledge-intensive tasks.

\section{Scientific Artifacts}
\label{sec:appendix_artifacts}
\subsection{Licensing}
We use publicly available datasets, models, and software artifacts in accordance with their respective licenses and terms of use. 

\subsection{Intended Use}
All external artifacts used in this work are employed strictly in accordance with their intended usage and licensing terms. Furthermore, any derivatives or findings from this study are designated solely for academic research purposes.

\subsection{Data Privacy and Safety}
All datasets used in this work are widely standard academic benchmarks that have already been vetted for privacy and toxicity. We manually inspected random samples of the data and confirmed that they do not contain personally identifying information (PII) or offensive content.

\subsection{Documentation of Artifacts}
The evaluation benchmarks utilized in our experiments are standard, well-documented open-source datasets. Their text domains predominantly cover open-domain Wikipedia articles and general-knowledge text. All evaluation materials are restricted to standard English, focus purely on factual and linguistic reasoning, and do not involve specific demographic or sensitive user profiles.
\section{Use Of AI Assistants} 
We used generative AI tools, including ChatGPT, to assist with identifying grammatical and syntactic issues in the manuscript.

\begin{table*}[t]
\centering
\caption{Case study illustrating spurious correctness caused by evidence fabrication. Search-r1 reaches the correct answer despite failing to retrieve the key supporting evidence. Instead of grounding the prediction in retrieved information, the model fills the missing evidence using internal knowledge, producing a correct answer by chance. This hidden retrieval failure may introduce misleading reward signals during reinforcement learning optimization.}
\label{tab:case_study_1}
\small
\begin{tabularx}{0.95\linewidth}{X}
\toprule
\textbf{Question:} Where was the \textit{Louisiana Purchase} signed in 1803? \\
\midrule

\textbf{Search-r1 (\textcolor{teal}{Success})} \\
\addlinespace[0.3em]
\texttt{<think>} To answer this question, I first need to find the location where the \textit{Louisiana Purchase} was signed in 1803. \texttt{</think>} \\
\texttt{<search>} where was the Louisiana Purchase signed in 1803 \texttt{</search>} \\
\texttt{<information>} \\
Doc1: Background of the purchase (\textit{ the U.S. acquired the Louisiana territory from France in 1803 for about \$15 million.}) \\
Doc2: States the treaty was signed on April 30, 1803, and that documents arrived in Washington D.C. on July 14. (\textit{no mention of the signing location.}) \\
Doc3: Covers Senate ratification on October 20, 1803, and transfer ceremonies held in New Orleans and St. Louis.( \textit{no mention of Paris.} ) \\
\texttt{</information>} \\
\texttt{<think>} From the search results, I know that the Louisiana Purchase was signed in Paris, France.(\textit{Paris never appears in any retrieved document. The model asserts the location without grounding it in the retrieval results.}) \texttt{</think>} \\
\textbf{Answer:} \textcolor{teal}{Paris}(\textit{correctness by chance}) \\

\bottomrule
\end{tabularx}
\end{table*}

\begin{table*}[t]
\centering
\caption{Case study comparing Search-r1 and SOAR on a multi-hop question. Search-r1 prematurely relies on hallucinated internal knowledge, leading to insufficient evidence retrieval and an incorrect answer. In contrast, SOAR explicitly separates internal and external knowledge sources, detects source inconsistency, and triggers deeper evidence retrieval, enabling accurate reasoning and answer generation.}
\label{tab:case_study_2}
\small
\begin{tabularx}{0.95\linewidth}{X}
\toprule
\textbf{Question:} Which film was released earlier, \textit{Ek Ladka Ek Ladki} or \textit{The Original Latin Kings Of Comedy}? \\
\midrule

\textbf{Search-r1 (\textcolor{red}{Failure})} \\
\addlinespace[0.3em]
\texttt{<think>} First, I need to determine the release dates of both films. \texttt{</think>} \\
\texttt{<search>} which film was released earlier, Ek Ladka Ek Ladki or The Original Latin Kings Of Comedy \texttt{</search>} \\
\texttt{<information>} \\
...\textit{Ek Ladka Ek Ladki} is an Indian Hindi-language romantic comedy film released on June 19, 1992... \\
\textit{\textbf{(No information about \textit{The Original Latin Kings Of Comedy})}} \\
\texttt{</information>} \\
\texttt{<think>} I found that \textit{Ek Ladka Ek Ladki} was released in 1992, while \textit{The Original Latin Kings Of Comedy} was released in \textcolor{red}{1978}. \textit{\textbf{(Hallucination)}} \texttt{</think>} \\
\textbf{Answer:} \textcolor{red}{The Original Latin Kings Of Comedy} \\

\midrule

\textbf{SOAR (\textcolor{teal}{Success})} \\
\addlinespace[0.3em]
\texttt{<think>} To answer this question, I need to find the specific release dates for both films. \texttt{</think>} \\
\texttt{<search>} Which film was released earlier, Ek Ladka Ek Ladki or The Original Latin Kings Of Comedy? \texttt{</search>} \\
\texttt{<information>} \\
\textbf{Internal:} \par ...Original Latin Kings Of Comedy was released in 1978... \par  \textbf{External:} \par ...\textit{Ek Ladka Ek Ladki} is an Indian Hindi-language romantic comedy film released on June 19, 1992... \\ 
\textit{\textbf{(No information about \textit{The Original Latin Kings Of Comedy})}} \\
\texttt{</information>} \\

\texttt{<think>} \textit{Ek Ladka Ek Ladki} was released in 1992. \textcolor{teal}{However, the internal date for the second film conflicts with knowledge and requires external verification.} \texttt{</think>} \\
\texttt{<search>} When was The Original Latin Kings Of Comedy released? \texttt{</search>} \\
\texttt{<information>} \\
\textbf{Internal:} \par ...Original Latin Kings Of Comedy was released in 1978... \par 
\textbf{External:} \par ...The Original Latin Kings of Comedy is a stand-up comedy film released in 2002... \\
\texttt{</information>} \\

\texttt{<think>} According to the external information, The Original Latin Kings Of Comedy was released in 2002. Comparing the two confirmed dates (1992 vs. 2002), \textit{Ek Ladka Ek Ladki} was released earlier. \texttt{</think>} \\
\textbf{Answer:} \textcolor{teal}{Ek Ladka Ek Ladki} \\
\bottomrule
\end{tabularx}
\end{table*}

\end{document}